\documentclass[11pt]{article}

\usepackage{acl}

\usepackage{times}
\usepackage{latexsym}

\usepackage[T1]{fontenc}

\usepackage[utf8]{inputenc}

\usepackage{microtype}

\usepackage{inconsolata}

\usepackage{graphicx}
\usepackage{multirow}
\usepackage{newunicodechar}
\newunicodechar{ʿ}{'}
\usepackage{amsmath}
\usepackage{amssymb}

\title{CVPD at QIAS 2025 Shared Task: An Efficient Encoder-Based Approach for Islamic Inheritance Reasoning}

\usepackage{authblk}

\author[1]{Salah Eddine Bekhouche}
\author[2]{Abdellah Zakaria Sellam}
\author[3]{Hichem Telli}
\author[2]{Cosimo Distante}
\author[4]{Abdenour Hadid}

\affil[1]{University of the Basque Country UPV/EHU, San Sebastian, Spain}
\affil[2]{Institute of Applied Sciences and Intelligent Systems – CNR, Lecce, Italy}
\affil[3]{Laboratory of LESIA, University of Biskra, Algeria}
\affil[4]{Sorbonne University Abu Dhabi, UAE}

\begin{document}
\maketitle
\begin{abstract}
Islamic inheritance law (\textit{ʿIlm al-Mawārīth}) requires precise identification of heirs and calculation of shares, which poses a challenge for AI. In this paper, we present a lightweight framework for solving multiple-choice inheritance questions using a specialised Arabic text encoder and Attentive Relevance Scoring (ARS). The system ranks answer options according to semantic relevance, and enables fast, on-device inference without generative reasoning. We evaluate Arabic encoders (MARBERT, ArabicBERT, AraBERT) and compare them with API-based LLMs (Gemini, DeepSeek) on the QIAS 2025 dataset. While large models achieve an accuracy of up to 87.6\%, they require more resources and are context-dependent. Our MARBERT-based approach achieves 69.87\% accuracy, presenting a compelling case for efficiency, on-device deployability, and privacy. While this is lower than the 87.6\% achieved by the best-performing LLM, our work quantifies a critical trade-off between the peak performance of large models and the practical advantages of smaller, specialized systems in high-stakes domains.
\end{abstract}
\section{Introduction}
\label{sec:introduction}
Large Language Models (LLMs) such as GPT-4 \cite{openai2023}, Gemini \cite{gemini2023}, and Deepseek-v3 \cite{liu2024deepseek} have advanced natural language processing, and show strong reasoning capabilities on many topics. However, as they were mainly trained on general web data, they often struggle in specialised domains with high accuracy \cite{bubeck2023sparks}. Islamic inheritance law (\textit{ʿIlm al-Mawārīth}) is one such area, which is based on fixed rules from the Qur'an and Sunnah and requires a precise understanding of the law and accurate mathematical proportion calculations \cite{esmaeili2012islamic, phillips1995treatise}. The complexity arises from rules such as \textit{far\={a}\textquotesingle{}i\d{d}} (fixed shares), \textit{ʿawl} (reduction of shares if more than one), and \textit{radd} (increase of shares if less than one) \cite{el-far_2011_islamic}, where errors can cause serious legal and financial problems. General LLMs often fail at such tasks due to the multi-step reasoning and strict numerical precision, especially in Arabic contexts \cite{arabi2023large}. Reinforcing this point, a recent comprehensive study by \cite{bouchekif2025islamic} specifically assessed LLMs on Islamic inheritance law, providing empirical evidence of their limitations in this domain. Therefore, the \textit{QIAS 2025} SubTask 1 becomes a valuable benchmark for the assessment \cite{qias2025}.
This paper presents a lightweight framework developed for Islamic inheritance reasoning to address these challenges. Our approach combines a pre-trained Arabic text encoder with an Attentive Relevance Scoring (ARS) module. Instead of generating step-by-step generative answers, the system measures how strongly each possible answer relates to the question. The ARS module then ranks the options and selects the correct legal and mathematical outcome. This design focuses on accuracy and efficiency, providing a more feasible solution than large LLMs requiring high computational resources.
We compare our specialised model with several leading general-purpose LLMs, including Gemini and DeepSeek, using the official QIAS 2025 dataset. Our experiments show that large models are prone to certain types of errors, especially under specific inference conditions. Our targeted approach, while not perfect, presents an alternative with a different performance and error profile, prioritizing consistency and efficiency. The primary contributions of this work are threefold:
\begin{enumerate}
    \item We present an efficient, specialized framework that applies an attentive relevance scoring mechanism to pre-trained Arabic encoders for Islamic inheritance reasoning.
    \item We provide a comparative analysis comparing our specialized model with SOTA general-purpose LLMs, highlighting the significant impact of inference strategies (batched vs. single input) on LLM performance.
    \item We provide empirical evidence of the practical advantages (efficiency, privacy, deployability) of domain-specific models, offering a viable alternative to resource-intensive LLMs despite a performance trade-off.
\end{enumerate}The remainder of this paper is organised as follows: Section~\ref{sec:methodology} describes our approach in detail; Section~\ref{sec:results_discussion} presents results and discussion; and Section~\ref{sec:conclusion} concludes with future research directions.

\section{Methodology}
\label{sec:methodology}
\begin{figure}
    \centering
    \includegraphics[width=0.8\linewidth]{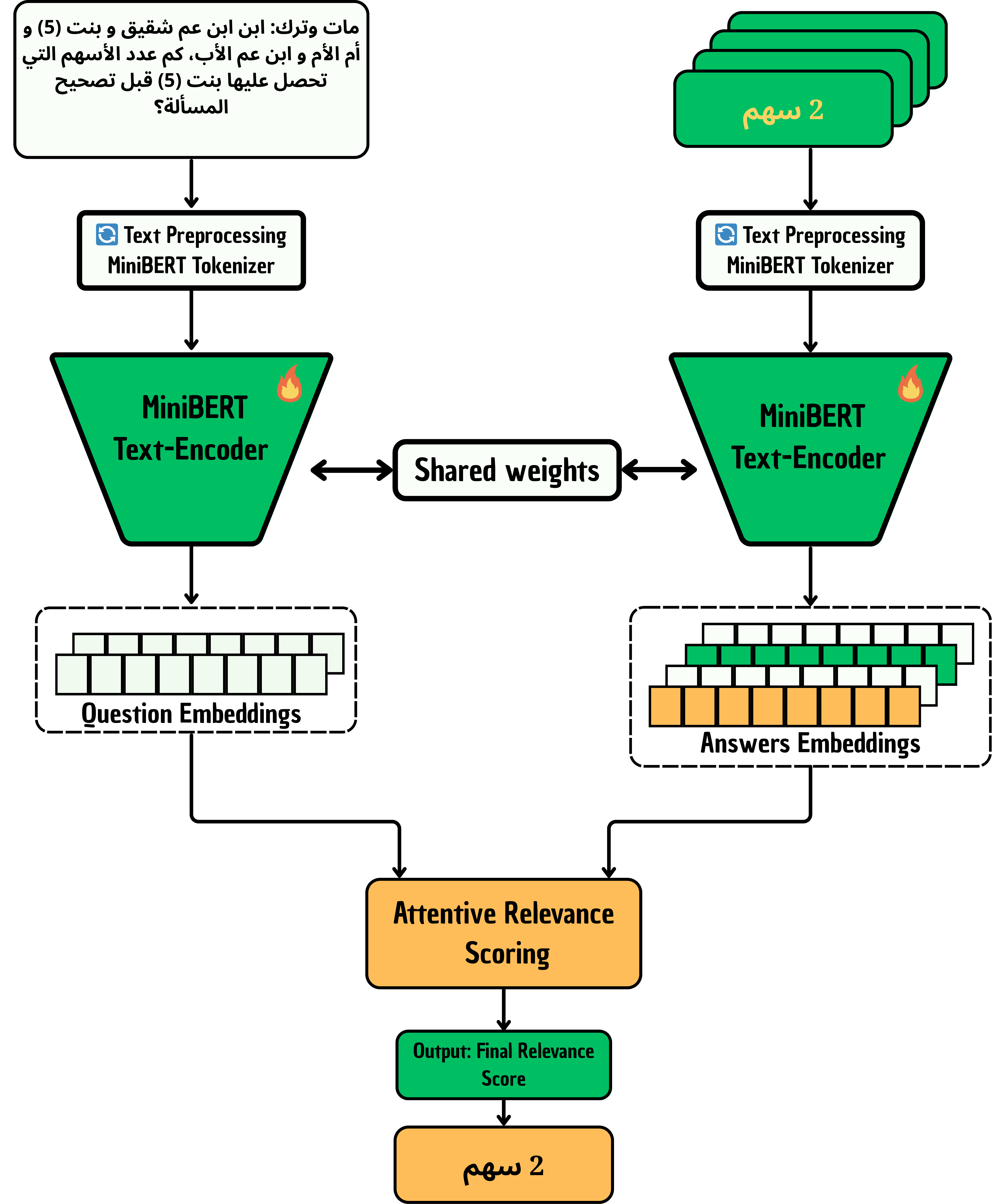}
    \caption{The proposed architecture. Parallel Text Encoders convert a question and answer into Question Embeddings and Answer Embeddings. An Attentive Relevance Scoring module then compares these embeddings to output a Final Relevance Score}
    \label{fig:general_architecture}
\end{figure}
This section describes our proposed hybrid architecture that combines an Arabic text encoder with a scoring mechanism called Attentive Relevance Scoring (ARS)~\cite{bekhouche2025enhanced}. We evaluate several Arabic text encoders in this setup. The method aims to improve question answering for Islamic inheritance law by capturing complex semantic relationships while keeping computation lightweight, making it suitable for low-resource environments and edge devices without cloud access.
A key design choice is that our approach does not rely on explicit reasoning. Instead, it focuses on providing a fast and low-cost inference solution directly on the device. The text encoder processes both the question and candidate answers, producing dense vector embeddings. The ARS module then scores each candidate answer by assigning higher weights to terms that are contextually important, enabling the system to capture fine-grained details in legal terminology.
As shown in Figure~\ref{fig:general_architecture}, the system operates in two stages: (1) the encoder generates semantic embeddings for the question and answers, and (2) ARS refines the ranking by computing a final relevance score. It is important to clarify that our approach does not perform explicit, step-by-step symbolic reasoning. Instead, it is designed to solve this complex reasoning task by learning to identify the candidate answer with the highest semantic relevance to the question.

\subsection{Text Encoder}
We experiment with five Arabic text encoders: ArabicBERT-Mini~\cite{safaya2020kuisail}, ArabicBERT~\cite{safaya2020kuisail}, AraBERT~\cite{antoun2020arabert}, MARBERT~\cite{abdul2020arbert}, and QARiB~\cite{abdelali2021pre}.  
Given a question $q$ and a candidate answer $c$, the encoder processes each independently, producing two types of representations: 
(1) \textbf{Sequence-level representations}, denoted as $\mathbf{H}_q \in \mathbb{R}^{B \times L \times d}$ and $\mathbf{H}_c \in \mathbb{R}^{B \times L \times d}$, which capture contextual embeddings for each token in the question and the answer.  
(2) \textbf{Pooled representations} from the final-layer [CLS] token.

Here, $B$ is the batch size, $L$ is the input length, and $d$ is the hidden dimension of the model. For all models, $L = 512$. The hidden size $d$ is $256$ for ArabicBERT-Mini and $768$ for the other encoders.
For global semantic representation, we extract the [CLS] token embedding from the final layer and apply $\ell_2$ normalization:
\begin{equation}
\label{eq:cls_embeddings}
\begin{aligned}
    \mathbf{q}_{\text{emb}} &= \text{Norm}(E(q)_{[\text{CLS}]}) \in \mathbb{R}^d, \\
    \mathbf{c}_{\text{emb}} &= \text{Norm}(E(c)_{[\text{CLS}]}) \in \mathbb{R}^d,
\end{aligned}
\end{equation}
where $\text{Norm}(\cdot)$ is $\ell_2$ normalization. This normalization projects embeddings onto the unit hypersphere, improving stability in similarity computations.
\subsection{Attentive Relevance Scoring}
The ARS module~\cite{bekhouche2025enhanced} computes adaptive semantic similarity between the question and candidate embeddings via a trainable interaction model.
First, both embeddings are projected into a shared latent space:
\begin{equation}
    \mathbf{h}_q = W_q \mathbf{q}_{\text{emb}}, \quad
    \mathbf{h}_c = W_c \mathbf{c}_{\text{emb}},
\end{equation}
where $W_q, W_c \in \mathbb{R}^{h \times d}$ are learnable projection matrices and $h$ is the shared hidden dimensionality.
Next, element-wise multiplication is applied, followed by a non-linear activation to compute the interaction vector $\mathbf{v}_{\text{int}}$:
\begin{equation}
\label{eq:interaction_vector}
    \mathbf{v}_{\text{int}} = \tanh(\mathbf{h}_q \odot \mathbf{h}_c),
\end{equation}
where $\odot$ denotes element-wise multiplication and $\tanh(\cdot)$ is the hyperbolic tangent function.
Finally, the relevance score $r$ is obtained using an attention vector $w_{\text{att}} \in \mathbb{R}^h$:
\begin{equation}
    r = \sigma\left( w_{\text{att}}^\top \mathbf{v}_{\text{int}} \right),
\end{equation}
where $\sigma(\cdot)$ is the sigmoid function.

\subsection{Training Objective}
To train the model effectively, we employ a composite training objective designed to optimize for both semantic representation and accurate ranking. This objective is composed of three distinct loss functions, each with a specific goal:

\begin{itemize}
    \item \textit{Contrastive Loss} ($\mathcal{L}_{\text{cons}}$): Aligns the embeddings of correct question-answer pairs while pushing them apart from incorrect pairs.
    \item \textit{Dynamic Relevance Loss} ($\mathcal{L}_{\text{dyn}}$): Directly supervises the final ARS scores to ensure the model produces confident and well-calibrated rankings.
    \item \textit{Relevance Score Logit Regularization} ($\mathcal{L}_{\text{reg}}$): Stabilizes training by encouraging variance in the pre-activation logits, preventing score collapse.
\end{itemize}

The total loss, $\mathcal{L}_{\text{total}}$, is a weighted sum of these components, formulated as:
\begin{equation}
\label{eq:total_loss}
\mathcal{L}_{\text{total}} = \alpha \mathcal{L}_{\text{cons}} + \beta \mathcal{L}_{\text{dyn}} + \gamma \mathcal{L}_{\text{reg}}
\end{equation}
We empirically set the balancing weights to $\alpha=0.4$, $\beta=0.4$, and $\gamma=0.2$. A detailed mathematical formulation for each component is provided in Appendix~\ref{app:loss_functions}.

\section{Results and Discussion}
\label{sec:results_discussion}

\subsection{Dataset}

The dataset in this study is from the official release of SubTask 1: Islamic Inheritance Reasoning in the QIAS 2025 challenge. It covers the rule-based field of Islamic inheritance law, where systems must understand scenarios, identify heirs, apply fixed-share rules, handle diminution and radd return, and calculate exact shares. All questions are multiple-choice with one correct answer, grouped into Beginner, Intermediate, and Advanced levels. The training set has 9,446 samples (5,095 Beginner, 3,431 Intermediate, 920 Advanced), the validation set has 1,000 samples (500 Beginner, 300 Intermediate, 200 Advanced), and the test set has 1,000 samples (500 Beginner, 500 Advanced, no labels). Training and validation have six labels (A–F), with C most common; the test set is unlabeled. Beginner questions involve simple share identification, Intermediate include adjusted shares after radd, and Advanced require full monetary distribution. This dataset is well-suited for testing both language understanding and precise numerical reasoning in Islamic law.

\subsection{Experimental Setup}
Experiments were performed on a system with seven NVIDIA L4 GPUs, each with 24 GB of VRAM, using a distributed multi-GPU training strategy. Mixed-precision training was not used, and the gradient accumulation step was set to 1 for stability.
Optimization was done with the AdamW optimizer, starting at \(1 \times 10^{-4}\) and \(\epsilon = 1 \times 10^{-8}\). A cosine annealing scheduler was employed to adjust the learning rate, which was warmed up to 10\% of its target before decaying. Gradient clipping with a maximum norm of 0.5 was applied for numerical stability.

\begin{table*}[]
\centering
\begin{tabular}{|l|l|l|ll|}
\hline
\multirow{2}{*}{Model}                                        & \multirow{2}{*}{Params (M) $\downarrow$} & \multirow{2}{*}{GFlops $\downarrow$} & \multicolumn{2}{l|}{Results}      \\ \cline{4-5} 
                                                              &                             &                         & \multicolumn{1}{l|}{Valid $\uparrow$} & Test $\uparrow$ \\ \hline
ArabicBERT-Mini \cite{safaya2020kuisail} + ARS  & 11.6                & 10.3             & \multicolumn{1}{l|}{65.62\%}      & 64.23\%      \\ \hline
ArabicBERT \cite{safaya2020kuisail} + ARS   & 110.7                       & 71.1                    & \multicolumn{1}{l|}{69.08\%}      & 67.19\%      \\ \hline
AraBERT \cite{antoun2020arabert}  + ARS            & 135.3                       & 96.2                    & \multicolumn{1}{l|}{73.85\%}      & 68.46\%      \\ \hline
MARBERT \cite{abdul2020arbert} + ARS & 162.9                       & 124.5                   & \multicolumn{1}{l|}{\textbf{77.32\%}}      & \textbf{69.87\%}      \\ \hline
QARiB \cite{abdelali2021pre} + ARS         & 135.3                       & 96.2                    & \multicolumn{1}{l|}{74.18\%}      & 68.63\%      \\ \hline
\end{tabular}
\caption{Performance and computational cost of various Arabic text encoders within our proposed framework on the QIAS 2025 SubTask 1 validation and test sets. MARBERT achieves the highest accuracy, demonstrating its superior ability to handle the linguistic nuances of Islamic inheritance law. Bold values indicate the best performance in each column.}
\label{tab:encoder_comparison}
\end{table*}

\subsection{Results and Discussion}
Table~\ref{tab:encoder_comparison} summarizes the performance and computational costs of various Arabic text encoders in our framework. MARBERT achieved the highest validation and test sets accuracy, showcasing its strong ability to capture the linguistic and domain-specific nuances needed for Islamic inheritance reasoning. Previous research supports that MARBERT, which is trained on extensive Arabic social media data, effectively handles complex morphology and semantic variations.
While this analysis primarily compares our models with state-of-the-art (SOTA) large language models (LLMs), future work should benchmark against traditional non-neural baselines (e.g., TF-IDF with cosine similarity) to quantify the advantages of deep learning methods, especially for lower-parameter encoders.
We also tested API-based LLMs using two inference strategies to assess the impact of context size on performance. The primary method involved a batched approach with 50 questions in a single prompt, which proved efficient but created a large context window. By contrast, the single-question method (used for testing Gemini-2.5-flash) improved accuracy significantly, from 68.65\% to 87.60\%. This indicates that larger context windows can lead to errors due to cross-question interference.
Although API-based models like Gemini and DeepSeek variants outperform our locally trained models regarding accuracy, their high computational requirements prevent direct deployment on edge devices. While running them through cloud services is viable, it entails recurring costs, latency issues, and privacy concerns, making local solutions more attractive in constrained or sensitive environments.
Ultimately, these findings reveal a trade-off between performance and deployability. A model with around 70\% accuracy is best suited as an assistive tool for legal experts rather than an autonomous decision-maker, facilitating rapid analysis or verification of simple cases, while human oversight remains essential. This positions such models as efficient assistants for on-device or offline scenarios where cloud access is not feasible. Additionally, our experiments demonstrate that inference setup and input structuring significantly impact model behavior, highlighting the importance of evaluation settings when comparing LLM-based systems.

\begin{table}[h]
\centering
\begin{tabular}{lcc}
\hline
\textbf{Base Model} & \textbf{Reasoning} & \textbf{ACC $\uparrow$} \\
\hline
deepseek-chat      & No  & 66.40\%  \\
deepseek-reasoner  & Yes & 69.40\%  \\
gemini-2.0-flash   & No  & 60.44\%  \\
gemini-2.5-flash   & Yes & 68.65\%  \\
\hline
gemini-2.5-flash*   & Yes & 87.60\%  \\
\hline
\end{tabular}
\caption{Performance of API-based LLMs. All models were evaluated using a batched input of 50 questions, except where noted by an asterisk (*).}
\label{tab:model_reasoning_acc}
\end{table}
\section{Conclusion}
\label{sec:conclusion}

We presented a lightweight framework for automated Islamic inheritance reasoning (\textit{ʿIlm al-Mawārīth}), combining a specialized Arabic text encoder with an Attentive Relevance Scoring (ARS) mechanism for multiple-choice questions. Our local model, using MARBERT, achieved a test accuracy of 69.87\%, which, while promising, is notably lower than the 87.60\% reached by leading API-based LLMs like Gemini. This performance difference stems from our model's core design, which forgoes explicit, step-by-step symbolic reasoning in favor of efficient semantic matching. Despite this accuracy trade-off, our approach offers significant advantages in computational efficiency, on-device deployability, and data privacy, making it a viable solution for resource-constrained or offline applications.

These results highlight a critical trade-off between peak performance and practical usability. The current accuracy level positions our system as a valuable \textbf{assistive tool} for legal experts rather than a fully autonomous decision-maker, underscoring the necessity of human oversight in such high-stakes, rule-based domains. This demonstrates that lightweight, domain-adapted models remain highly relevant for specific use cases. Future work will directly aim to close the accuracy gap by integrating symbolic reasoning capabilities to handle the precise calculations inherent in inheritance law. We will also explore hybrid approaches that combine the efficiency of our lightweight model with the reasoning power of large models to achieve an optimal balance of performance and practicality.

\section*{Acknowledgement} The authors thank Mr. Arturo Argentieri from CNR-ISASI Italy for his technical contribution to the multi-GPU computing facilities. This research was partially funded by the Italian Ministry of University and Research (MUR) with the project "Future Artificial Intelligence Research —FAIR"  Grant number PE0000013 CUP B53C22003630006.




\bibliography{custom}

\appendix

\section{Detailed Training Objective}
\label{app:loss_functions}

This section provides the detailed mathematical formulation of the three loss components used in our training objective. The total loss is defined as:
\begin{equation}
\mathcal{L}_{\text{total}} = 0.4 \cdot \mathcal{L}_{\text{cons}} + 0.4 \cdot \mathcal{L}_{\text{dyn}} + 0.2 \cdot \mathcal{L}_{\text{reg}}
\end{equation}

\subsection{Contrastive Loss ($\mathcal{L}_{\text{cons}}$)}
We use an InfoNCE-based contrastive loss on the \texttt{[CLS]} token embeddings. This loss aims to pull the question embedding ($\mathbf{q}$) closer to the correct answer embedding ($\mathbf{c}^+$) and push it away from the five incorrect answer embeddings ($\mathbf{c}^-$).
\begin{equation}
    \mathcal{L}_{\text{cons}} = -\frac{1}{B} \sum_{i=1}^{B} \log\left( \frac{ e^{\text{sim}(\mathbf{q}_i, \mathbf{c}_i^+)} }{ e^{\text{sim}(\mathbf{q}_i, \mathbf{c}_i^+)} + \sum_{j=1}^{5} e^{\text{sim}(\mathbf{q}_i, \mathbf{c}_{i,j}^-)} } \right)
\end{equation}
where $\text{sim}(\mathbf{a}, \mathbf{b}) = (\mathbf{a}^\top \mathbf{b}) / \tau$. Here, $\mathbf{q}_i$ and $\mathbf{c}_i$ are the embeddings for the question and answers, and $\tau$ is a trainable temperature parameter.

\subsection{Dynamic Relevance Loss ($\mathcal{L}_{\text{dyn}}$)}
This loss directly supervises the final ARS scores ($r$) to ensure they are well-calibrated. It maximizes the score for the correct answer and minimizes the score for a randomly selected incorrect answer.
\begin{align}
    \mathcal{L}_{\text{dyn}} = -\frac{1}{B} \sum_{i=1}^{B} \left[ \log(r_i^+ + \epsilon) + \log(1 - r_i^- + \epsilon) \right]
\end{align}
Here, $r_i^+$ and $r_i^-$ are the sigmoid-activated ARS scores for the correct and a randomly chosen incorrect answer. The constant $\epsilon$ ensures numerical stability.

\subsection{Relevance Score Logit Regularization ($\mathcal{L}_{\text{reg}}$)}
To improve training stability, we apply a regularization loss on the raw, pre-sigmoid relevance scores (logits, $s$). This loss maximizes the variance of the logits within a batch, encouraging the model to use a wider dynamic range for its scores.
\begin{equation}
    \mathcal{L}_{\text{reg}} = -(\operatorname{Std}(s_{\text{batch}}^+) + \operatorname{Std}(s_{\text{batch}}^-))
\end{equation}
where $s_{\text{batch}}^+$ and $s_{\text{batch}}^-$ are the sets of logits for all correct and incorrect answers across the batch. We minimize the negative standard deviation, which is equivalent to maximizing the standard deviation.

\end{document}